\documentclass[10pt, conference, compsocconf]{IEEEtran}
\usepackage{cite}
\usepackage[font=small,skip=0pt]{caption}
\usepackage{amsmath,amssymb,amsfonts}
\usepackage{algorithmic}
\usepackage{graphicx}
\usepackage{booktabs}
\usepackage{tabularx}
\usepackage{subcaption}
\usepackage{textcomp}
\usepackage{xcolor}
\usepackage{multirow}
\usepackage[utf8]{inputenc}
\usepackage{amsmath}
\usepackage{bm}
\usepackage{xfrac}
\usepackage{numprint}
\usepackage{url}
\usepackage{array}
\usepackage{etoolbox}
\usepackage{hyperref}

\newtoggle{blinded}
\togglefalse{blinded}

\newcommand{\OCRName}{Ours}

\npthousandsep{\,}

\def\BibTeX{{\rm B\kern-.05em{\sc i\kern-.025em b}\kern-.08em
    T\kern-.1667em\lower.7ex\hbox{E}\kern-.125emX}}

\begin{document}
\bstctlcite{IEEEexample:BSTcontrol}

\title{Efficient, Lexicon-Free OCR using Deep Learning
}

\iftoggle{blinded}{}{
\author{\IEEEauthorblockN{Marcin Namysl}
\IEEEauthorblockA{\textit{Fraunhofer IAIS} \\
53757 Sankt Augustin, Germany \\
Marcin.Namysl@iais.fraunhofer.de }
\and
\IEEEauthorblockN{Iuliu Konya}
\IEEEauthorblockA{\textit{Fraunhofer IAIS} \\
53757 Sankt Augustin, Germany \\
Iuliu.Konya@iais.fraunhofer.de }
}
}

\maketitle

\begin{abstract}
Contrary to popular belief, Optical Character Recognition (OCR) remains a challenging problem when text occurs in unconstrained environments, like natural scenes, due to geometrical distortions, complex backgrounds, and diverse fonts.
In this paper, we present a segmentation-free OCR system that combines deep learning methods, synthetic training data generation, and data augmentation techniques. 
We render synthetic training data using large text corpora and over \numprint{2000} fonts. 
To simulate text occurring in complex natural scenes, we augment extracted samples with geometric distortions and with a proposed data augmentation technique -- alpha-compositing with background textures.
Our models employ a convolutional neural network encoder to extract features from text images. 
Inspired by the recent progress in neural machine translation and language modeling, we examine the capabilities of both recurrent and convolutional neural networks in modeling the interactions between input elements. 
The proposed OCR system surpasses the accuracy of leading commercial and open-source engines on distorted text samples. 

\end{abstract}


\section{Introduction}
\label{sec:intro}
Optical character recognition (OCR) is one of the most widely studied problems in the field of pattern recognition and computer vision. 
It is not limited to printed but also handwritten documents~\cite{Graves2008}, as well as natural scene text~\cite{Shi2017}. The accuracy of various OCR methods has recently greatly improved due to advances in deep learning~\cite{ioffe2015,He2016,hu2018senet}. Moreover, many current open-source and commercial products reach a high recognition accuracy and good throughput for run-of-the-mill printed document images. While this has lead the research community to regard OCR as a largely solved problem,
we show that even the most successful and widespread OCR solutions are neither able to robustly handle  large font varieties, nor distorted texts, potentially superimposed on complex backgrounds. Such unconstrained environments for digital documents have already become predominant, due to the wide availability of mobile phones and various specialized video recording devices. 

In contrast to popular OCR engines, methods used in scene text recognition~
\cite{Lyu_2018_ECCV,Busta2017DeepTA} exploit computationally expensive network models, aiming to achieve the best possible recognition rates on popular benchmarks. Such methods are tuned to deal with significantly smaller amounts of text per image and are often constrained to predefined lexicons. Commonly used evaluation protocols substantially limit the diversity of symbols to be recognized, e.g., by ignoring all non-alphanumeric characters, or neglecting case sensitivity~\cite{Wang2011}.
Hence, models designed for scene text are generally inadequate for printed document OCR, whereas high throughput and support for great varieties of symbols are essential.

In this paper, we address the general OCR problem and try to overcome the limitations of both printed- and scene text recognition systems. To this end, we present a fast and robust deep learning multi-font OCR engine, which currently recognizes $132$ different character classes. Our models are trained almost exclusively using synthetically generated documents. 
We employ segmentation-free text recognition methods that require a much lower data labeling effort, making the resulting framework more readily extensible for new languages and scripts. Subsequently we propose a novel data augmentation technique that improves the robustness of neural models for text recognition. Several large and challenging datasets, consisting of both real and synthetically rendered documents, are used to evaluate all OCR methods. The comparison with leading established commercial (ABBYY FineReader~\footnote{\url{https://www.abbyy.com/en-eu/ocr-sdk/}}, OmniPage Capture~\footnote{\url{https://www.nuance.com/print-capture-and-pdf-solutions/optical-character-recognition/omnipage/omnipage-for-developers.html}}) and open-source engines (Tesseract~3~\cite{Smith2007}, Tesseract~4~\footnote{\url{https://github.com/tesseract-ocr/tesseract/wiki/4.0-with-LSTM}}) shows that the proposed solutions obtain significantly better recognition results with comparable execution time.

The remaining part of this paper is organized as follows: in Section~\ref{sec:related_work}, we highlight related research papers, while in Section~\ref{sec:data}, we describe the datasets used in our experiments, as well as the data augmentation routines. In Section~\ref{sec:arch}, we present the detailed system architecture, which is then evaluated and compared against several state-of-the-art OCR engines in Section~\ref{sec:eval}. Our conclusions, alongside a few worthwhile avenues for further investigations, are the subject of the final Section~\ref{sec:conclusions}.

\section{Related work}
\label{sec:related_work}
In this section, we review related approaches for printed-, handwritten- and scene text recognition. These can be broadly categorized into segmentation-based and segmentation-free methods.

Segmentation-based OCR methods recognize individual character hypotheses, explicitly or implicitly generated by a character segmentation method. The output is a recognition lattice containing various segmentation and recognition alternatives weighted by the classifier. The lattice is then decoded, e.g., via a greedy or beam search method and the decoding process may also make use of an external language model or allow the incorporation of certain (lexicon) constraints.

The PhotoOCR system for text extraction from smartphone imagery, proposed by Bissacco et al.\cite{Bissacco2013}, is a representative example of a segmentation-based OCR method.
They used a deep neural network trained on extracted histogram of oriented gradient (HOG) features for character classification and incorporated a character-level language model into the score function.

The accuracy of segmentation-based methods heavily suffers from segmentation errors and the lack of context information wider than a single cropped character-candidate image during classification. Improper incorporation of an external language model or lexicon constraints can degrade accuracy\cite{Smith2011}.
While offering a high flexibility in the choice of segmenters, classifiers, and decoders, segmentation-based approaches require a similarly high effort in order to tune optimally for specific application scenarios. Moreover, the precise weighting of all involved hypotheses must be re-computed from scratch as soon as one component is updated (e.g., the language model), whereas the process of data labeling (e.g., at the character/pixel level) is usually a painstaking endeavor, with a high cost in terms of human annotation labor.

Segmentation-free OCR methods eliminate the need for pre-segmented inputs. Cropped words or entire text lines are usually geometrically normalized~(\ref{sec:data:geom_norm}) and then can be directly recognized. Previous works on segmentation-free OCR~\cite{rashid2012scanning} employed Hidden Markov Models (HMMs) to avoid the difficulties of segmentation-based techniques.
Most of the recently developed segmentation-free solutions employ recurrent and convolutional neural networks.


Multi-directional, multi-dimensional recurrent neural networks (MDRNNs) currently enjoy a high popularity among researchers in the field of handwriting recognition~\cite{Graves2008} because of their ability to attain state-of-the-art recognition rates. 
They generalize standard Recurrent Neural Networks (RNNs) by providing recurrent connections along all spatio-temporal dimensions, making them robust to local distortions along any combination of the respective input dimensions. Bidirectional RNNs, consisting of two hidden layers that traverse the input sequence in opposite spatial directions (i.e., left-to-right and right-to-left), connected to a single output layer, were found to be well-suited for both handwriting\cite{Graves2009} and printed text recognition\cite{Breuel2013}. In order to mitigate the vanishing/exploding gradient problem, most RNNs use Long Short-Term Memory (LSTM) units (or variants thereof) as building blocks. A noteworthy extension to the LSTM cells are the "peephole" LSTM units\cite{Gers2000}, where the multiplicative gates compute their activations at the current time step using in addition the activation of the memory cell from the previous time step.

MDRNNs are computationally much more expensive than their basic 1-D variant, both during training and inference. Because of this, they have been less frequently explored in the field of printed document OCR. Instead, in order to overcome the issue of sensitivity to stroke variations along the vertical axis, researchers have proposed different solutions. For example, Breuel et al.~\cite{Breuel2013} combined a standard 1-D LSTM network architecture with a text line normalization method for performing OCR of printed Latin and Fraktur scripts. In a similar manner, by normalizing the positions and baselines of letters, Yousefi et al.~\cite{yousefi2015comparison} achieved superior performance and faster convergence with a 1-D LSTM network over a 2-D variant for Arabic handwriting recognition.

An additional advantage of segmentation-free approaches is their inherent ability to work directly on grayscale or full-color images. This increases the robustness and accuracy of text recognition, as any information loss caused by a previously mandatory binarization step can be avoided. Asad et al.~\cite{Asad2016} applied the 1-D LSTM network directly to original, blurred document images and were able to obtain state-of-the-art recognition results.
The introduction of convolutional neural networks (CNNs) allowed for a further jump in the recognition rates. 
Since CNNs are able to extract latent representations of input images, thus increasing robustness to local distortions, they can be successfully employed as a substitute for MD-LSTM layers. Breuel et al.~\cite{Breuel2017} proposed a model that combined CNNs and LSTMs for printed text recognition. Features extracted by CNNs were combined and fed into the LSTM network with a Connectionist Temporal Classification (CTC)\cite{Graves2006} output layer. 
A few recent methods have completely forgone the use of the computationally expensive recurrent layers and rely purely on convolutional layers for modeling the local context. 
Borisyuk et al.~\cite{Borisyuk2018} presented a scalable OCR system called \textit{Rosetta}, which employs a fully convolutional network (FCN) model followed by the CTC layer in order to extract the text depicted on input images uploaded daily at Facebook.



In the current work, we build upon the previously mentioned techniques and propose an end-to-end segmentation-free OCR system. Our approach is purely data-driven and can be adapted with minimal manual effort to different languages and scripts. Feature extraction from text images is realized using convolutional layers. Using the extracted features, we analyze the ability to model local context with both recurrent and fully convolutional sequence-to-sequence architectures. The alignment of the extracted features with ground-truth transcripts is realized via a CTC layer. To the best of our knowledge, this is the first work that compares fully convolutional and recurrent models in the context of OCR. 




\section{Datasets and data preparation}
\label{sec:data}
To train and evaluate our OCR system we prepared several datasets, consisting of both real and synthetic documents. This section describes each in detail, as well as the preparation of training, validation, and test samples, data augmentation techniques, and the geometric normalization procedure.


We collected $\numprint{16}$ pages of scanned historical and recent German-language newspapers as well as $\numprint{11}$ contemporary German invoices. All documents were deskewed and pre-processed via document layout analysis algorithms, providing us with the geometrical and logical document structure, including bounding boxes, baseline positions, and x-height values for each text line. The initial transcriptions obtained using the Tesseract~\cite{Smith2007} OCR engine were manually corrected.


Even without the need for the character- or word-level ground truth, the manual annotation process proved to be error-prone and time-consuming. Motivated by the work of Jaderberg et al.~\cite{Jaderberg2014}, we developed an automatic synthetic data generation process. Two large text corpora, namely the English and German Wikipedia dump files\footnote{\url{https://dumps.wikimedia.org/}}, were used as training sources for generating sentences. For validation and test purposes, we used a corpus from the Leipzig Corpora Collection\cite{goldhahn2012building}. The texts were rendered using a set of over $\numprint{2000}$ serif, sans serif, and monospace fonts\footnote{\url{https://fonts.google.com/}}. 


The generation process first selects a piece of text (up to $40$ characters) from the corpus and renders it on an image with a randomly chosen font. The associated attributes (i.e., bounding boxes, baseline positions, and x-height values) used for rendering are stored in the corresponding document layout structure. A counter for the number of occurrences of every individual character in the generated dataset is maintained and used to guide the text extraction mechanism to choose text pieces containing the less frequently represented symbols. Upon generating enough text line samples to fill an image of pre-specified dimensions (e.g., $\numprint{3500}\times\numprint{5000}$ pixels), the image is saved on disk together with the associated layout information. The procedure described above is repeated until the number of occurrences of each symbol reaches a required minimum level ($\numprint{5000}$, $\numprint{100}$ and $\numprint{100}$ in our synthetic training, validation, and test set, respectively), which guarantees that even rare characters are sufficiently well represented in each of the generated datasets, or until all text files have been processed.
By using sentences from real corpora we ensure that the sampled character and n-gram distribution is the same as that of natural language texts. 

A summary of our data sources is presented in~\tablename~\ref{tab:datasources}. We train and recognize \numprint{132} different character classes, including basic lower and upper case Latin letters, whitespace character, German umlauts, ligature ß, digits, punctuation marks, subscripts, superscripts, as well as mathematical, currency and other commonly used symbols. The training data consists of about $\numprint{6.4}$ million characters, $95.9\%$ of which were synthetically generated.

{\setlength{\tabcolsep}{3pt}\renewcommand{\arraystretch}{1.0}
\newcommand\Tstrut{\rule{0pt}{2.2ex}}         
\begin{table}[htbp]
\caption{Summary of the used data sources.}
\label{tab:datasources}
\begin{center}
\begin{tabular}{lrrrrrr}
\toprule
\multirow{2}{*}{} & \multicolumn{2}{c}{\textbf{Newspapers}} & \multicolumn{1}{c}{\textbf{Invoices}} & \multicolumn{3}{c}{\textbf{Synthetic documents}}\\
\cmidrule(lr){2-3}
\cmidrule(lr){4-4}
\cmidrule(lr){5-7}
& Training & Test & Test & Training & Validation & Test\\
\midrule
Documents  & $\numprint{12}$     & $\numprint{4}$     & $\numprint{11}$    & $\numprint{390}$     & $\numprint{10}$     & $\numprint{9}$ \\
Text lines & $\numprint{7049}$   & $\numprint{1943}$  & $\numprint{486}$   & $\numprint{200030}$  & $\numprint{4827}$   & $\numprint{4650}$ \\
GT length  & $\numprint{260041}$ & $\numprint{66327}$ & $\numprint{15374}$ & $\numprint{6128394}$ & $\numprint{151653}$ & $\numprint{141280}$  \\
\bottomrule
\end{tabular}
\end{center}
\end{table}}

The batches containing the final training and validation samples are generated on the fly, as in the following. Text line images are randomly selected from the corresponding (training or validation) dataset and the associated layout information is used to normalize each sample. Note that the normalization step is optional (see also \ref{sec:data:geom_norm}), since especially in the case of scene text it may be too computationally expensive and error-prone to extract exact baselines and x-heights at inference time. All samples are re-scaled to a fixed height of $\numprint{32}$ pixels, while maintaining the aspect ratio. This particular choice for the sample height was determined experimentally. Larger sample heights did not improve recognition accuracy for skew-free text lines. However, if the targeted use case involves the recognition of relatively long, free-form text lines, the use of taller samples is warranted. Since text lines lengths vary greatly, the corresponding images must be (zero) padded appropriately to fit the widest element within the batch. We minimize the amount of padding by composing batches from text lines having similar widths. Subsequently, random data augmentation methods are dynamically applied to each sample (\ref{sec:data:augment}).


\begin{figure}[!htbp]
{\setlength{\tabcolsep}{0pt}\renewcommand{\arraystretch}{0.5}
\begin{center}
\begin{tabular}{c}
\includegraphics[width=1.0\columnwidth]{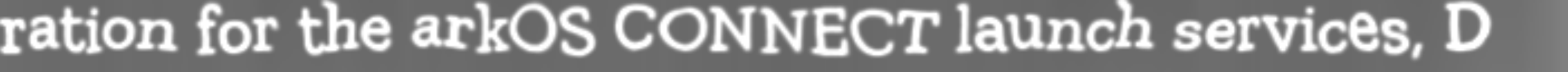}\\
\includegraphics[width=1.0\columnwidth]{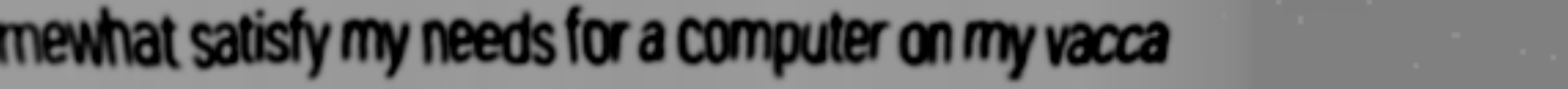}\\
\includegraphics[width=1.0\columnwidth]{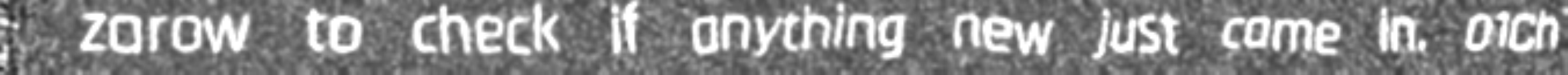}\\
\includegraphics[width=1.0\columnwidth]{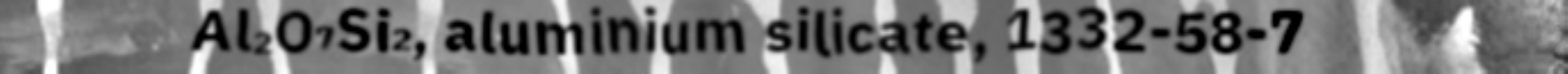}\\
\includegraphics[width=1.0\columnwidth]{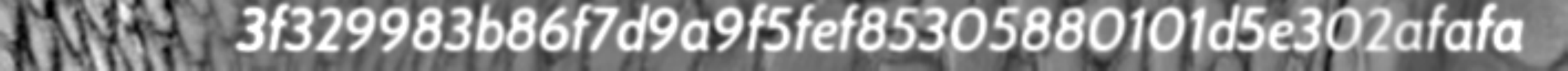}\\
\includegraphics[width=1.0\columnwidth]{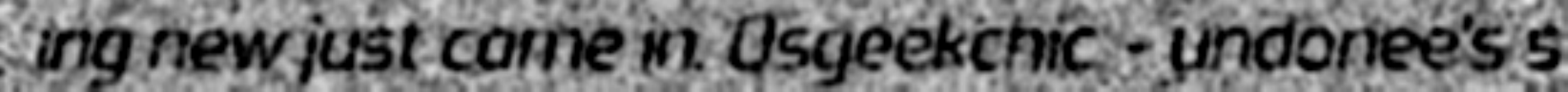}\\
\includegraphics[width=1.0\columnwidth]{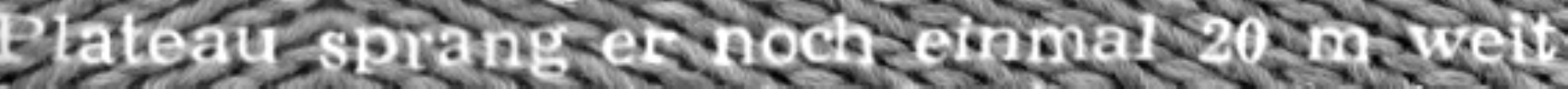}\\
\includegraphics[width=1.0\columnwidth]{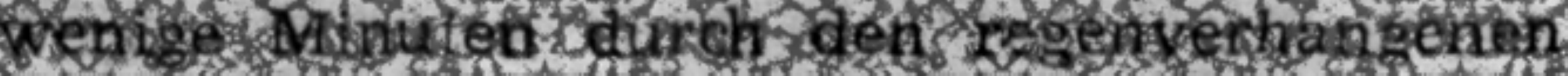}\\
\includegraphics[width=1.0\columnwidth]{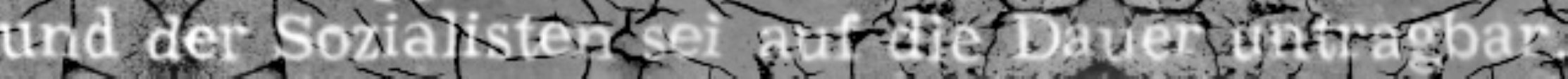}\\
\includegraphics[width=1.0\columnwidth]{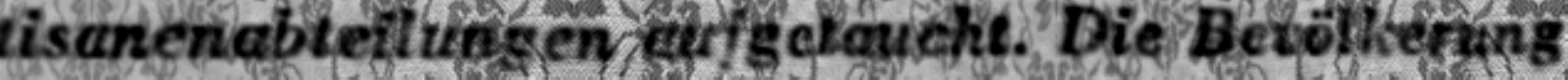}\\
\includegraphics[width=1.0\columnwidth]{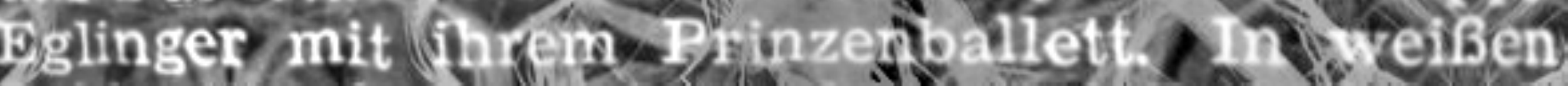}\\
\includegraphics[width=1.0\columnwidth]{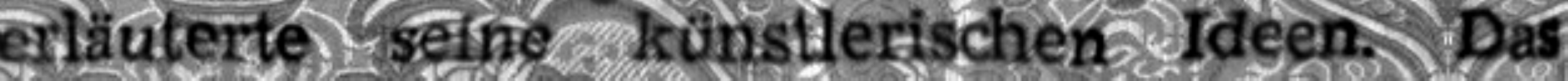}
\end{tabular}
\end{center}
}
\caption{Training and validation data samples from our system.}
\label{fig:samples}
\end{figure}

\subsection{Data augmentation}
\label{sec:data:augment}
We apply standard data augmentation methods, like Gaussian smoothing, perspective distortions, morphological filtering, downscaling, additive noise, and elastic distortions\cite{Simard2003} during training and validation.
Additionally, we propose a novel augmentation technique --- alpha compositing\cite{Porter1984} with background texture images. 
Each time a specific sample is presented to the network, it is alpha-composited with a randomly selected background texture image (\figurename~\ref{fig:samples}). 
By randomly altering backgrounds of training samples, the network is guided to focus on significant text features and learns to ignore background noise.
The techniques mentioned above are applied dynamically both to training and validation samples.
In contrast to the approach proposed by Jaderberg et al.~\cite{Jaderberg2014}, we render undistorted synthetic documents once, and then apply random data augmentations dynamically. This allows us to efficiently generate samples and eliminates the significant overhead caused by disk I/O operations.

\subsection{Geometric normalization}
\label{sec:data:geom_norm}
Breuel et al.~\cite{Breuel2017} recommended that text line images should be geometrically normalized prior to recognition. We trained models with and without such normalization in order to verify this assumption. The normalization step is performed per text line, before feature extraction. During training, we use the saved text line attributes, whereas at inference time, the layout analysis algorithm provides the baseline position and the x-height values for each line. Using the baseline information, the skew of the text lines is corrected. The scale of each image is normalized, so that the distance between the baseline and the x-height line, as well as the heights of ascenders and descenders, are approximately constant.

For the unnormalized case, the normalization procedure is skipped entirely and each cropped text line sample is further passed on to the feature extractor. This case is especially relevant for scene text images, where normalization information is usually unavailable and expensive to compute separately.

\section{System architecture}
\label{sec:arch}
\begin{figure}[!htbp]
\centering
\includegraphics[width=0.70\columnwidth]{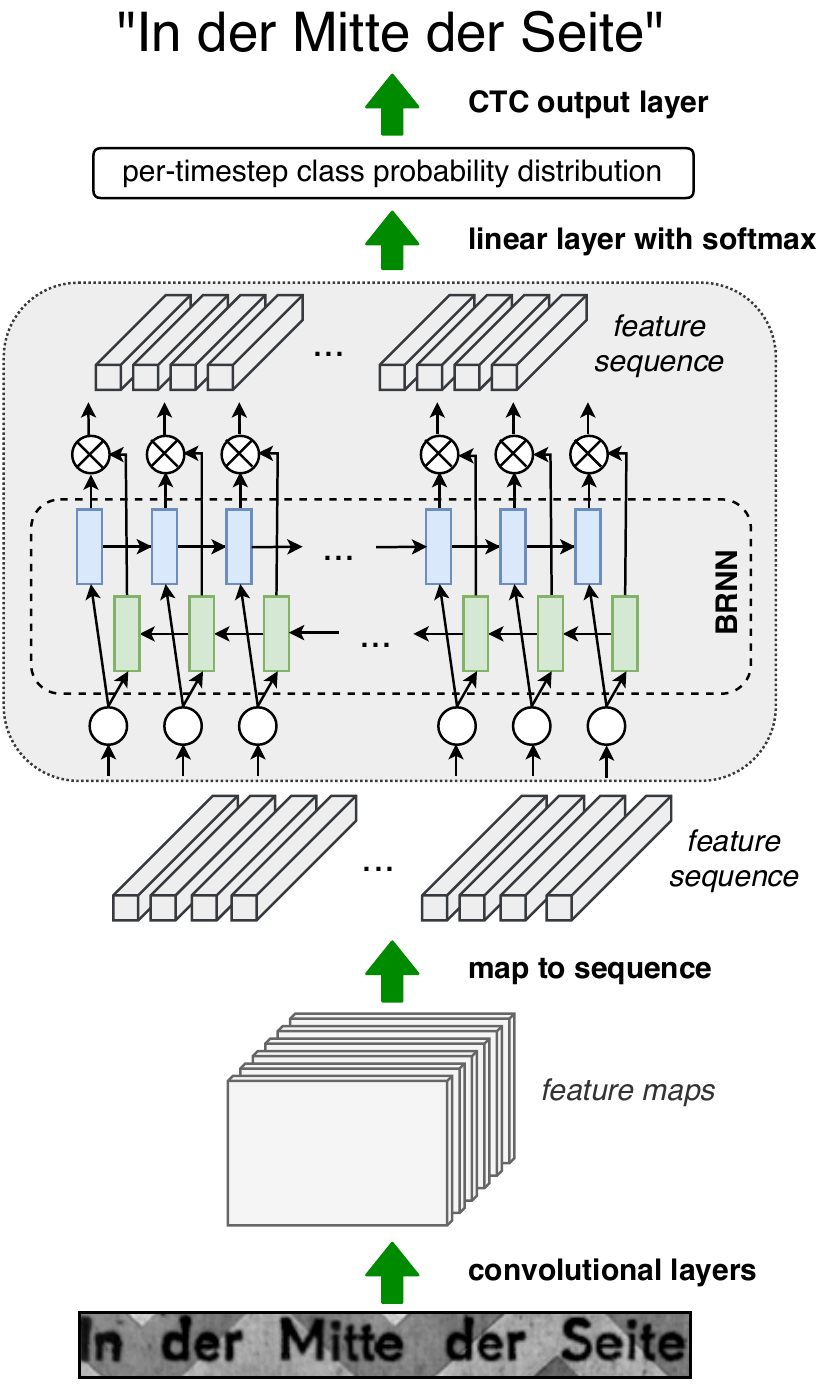}
\caption{The architecture of our system. The gray region indicates a recurrent block omitted in case of the fully convolutional model. }
\label{fig:arch}
\end{figure}

The architecture of our hybrid CNN-LSTM model is depicted in \figurename~\ref{fig:arch} and is inspired by the \textit{CRNN}~\cite{Shi2017} and \textit{Rosetta}~\cite{Borisyuk2018} systems. 
The bottom part consists of convolutional layers that extract high-level features of an image. Activation maps obtained by the last convolutional layer are transformed into a feature sequence with the \textit{map to sequence} operation. Specifically, 3D maps are sliced along their width dimension into 2D maps and then each map is flattened into a vector. 
The resulting feature sequence is fed to a bidirectional recurrent neural network with $\numprint{256}$ hidden units in both directions. The output sequences from both layers are concatenated and fed to a linear layer with softmax activation function to produce per-timestep probability distribution over the set of available classes.
The CTC output layer is employed to compute a loss between the network outputs and the ground truth transcriptions. During inference, CTC loss computation is replaced by greedy CTC decoding. 
\tablename~\ref{tab:cnn-lstm} presents a detailed structure of our recurrent model.

{\setlength{\tabcolsep}{8pt}\renewcommand{\arraystretch}{1.0}
\newcommand\Tstrut{\rule{0pt}{2.2ex}}         
\begin{table}[!htbp]
\caption{Detailed structure of our recurrent model.}
\label{tab:cnn-lstm}
\begin{center}
\begin{tabular}{lc}
\toprule
\textbf{Operation} & \textbf{Output volume size} \\
\midrule
Conv2d ($3{\times}3,64$; stride: $1{\times}1$) & $32{\times}\textit{W}{\times}64$\\
Max pooling ($2{\times}2$; stride: $2{\times}2$) & $16{\times}\sfrac{\textit{W}}{2}{\times}64$ \\
Conv2d ($3{\times}3,128$; stride: $1{\times}1$) & $16{\times}\sfrac{\textit{W}}{2}{\times}128$ \\
Max pooling ($2{\times}2$; stride: $2{\times}2$) & $8{\times}\sfrac{\textit{W}}{4}{\times}128$ \\
Map to sequence & $\sfrac{\textit{W}}{4}{\times}1024$ \\
Dropout (50\%) & --- \\
Bidirectional RNN (units: $2{\times}256$) & $\sfrac{\textit{W}}{4}{\times}512$ \\
Dropout (50\%) & --- \\
Linear mapping (units: \textit{num\_classes}) & $\sfrac{\textit{W}}{4}{\times}\textit{num\_classes}$ \\
CTC output layer & \textit{output\_sequence\_length} \\
\bottomrule
\end{tabular}
\end{center}
\end{table}}

In case of our fully convolutional model, feature sequences transformed by the \textit{map to sequence} operation (see the previous paragraph) are directly fed to a linear layer, skipping the recurrent components entirely. 
\tablename~\ref{tab:cnn-cnn} presents the detailed structure of our fully convolutional model.

{\setlength{\tabcolsep}{8pt}\renewcommand{\arraystretch}{1.0}
\newcommand\Tstrut{\rule{0pt}{2.2ex}}         
\begin{table}[!htbp]
\caption{Detailed structure of our fully convolutional model.}
\label{tab:cnn-cnn}
\begin{center}
\begin{tabular}{lc}
\toprule
\textbf{Operation} & \textbf{Output volume size}\\
\midrule
Conv2d ($7{\times}7,64$; stride: $2{\times}2$) & $16{\times}\sfrac{\textit{W}}{2}{\times}64$ \\
Max pooling ($2{\times}2$; stride: $2{\times}2$) & $8{\times}\sfrac{\textit{W}}{4}{\times}64$  \\
Conv2d ($3{\times}3,64$; stride: $1{\times}1$) & $8{\times}\sfrac{\textit{W}}{4}{\times}64$   \\
Conv2d ($3{\times}3,64$, stride: $1{\times}1$) & $8{\times}\sfrac{\textit{W}}{4}{\times}64$   \\
Conv2d ($3{\times}3,128$, stride: $2{\times}1$) & $4{\times}\sfrac{\textit{W}}{4}{\times}128$ \\
Conv2d ($3{\times}3,128$, stride: $1{\times}1$) & $4{\times}\sfrac{\textit{W}}{4}{\times}128$ \\
Conv2d ($3{\times}3,256$, stride: $2{\times}1$) & $2{\times}\sfrac{\textit{W}}{4}{\times}256$ \\
Conv2d ($3{\times}3,256$, stride: $1{\times}1$) & $2{\times}\sfrac{\textit{W}}{4}{\times}256$ \\
Conv2d ($3{\times}3,512$, stride: $2{\times}1$) & $1{\times}\sfrac{\textit{W}}{4}{\times}512$ \\
Conv2d ($3{\times}3,512$, stride: $1{\times}1$) & $1{\times}\sfrac{\textit{W}}{4}{\times}512$ \\
Conv2d ($3{\times}3,512$, stride: $1{\times}1$) & $1{\times}\sfrac{\textit{W}}{4}{\times}512$ \\
Map to sequence & $\sfrac{\textit{W}}{4}{\times}512$ \\
Linear layer (units: \textit{num\_classes}) & $\sfrac{\textit{W}}{4}{\times}\textit{num\_classes}$ \\
CTC output layer & \textit{output\_sequence\_length} \\
\bottomrule
\end{tabular}
\end{center}
\end{table}}{\

All models were trained via minibatch stochastic gradient descent using the Adaptive Moment Estimation (Adam) optimization method\cite{kingma2014adam}. 
The learning rate is decayed by a factor of $0.99$ every $\numprint{1000}$ iterations and has an initial value of $0.0006$ for the recurrent- and $0.001$ for the fully convolutional model. 
Batch normalization\cite{ioffe2015} is applied after every convolutional block to speed up the training. The hybrid models were trained for approximately $300$ epochs and the fully convolutional models for about $500$ epochs.

The Python interface of the Tensorflow~\cite{Abadi2016} framework was used for training all models. The inference timings were done via Tensorflow's C++ interface. 

\section{Evaluation and discussion}
\label{sec:eval}
We compare performance of our system with two established commercial OCR products: ABBYY FineReader~12 and OmniPage Capture SDK 20.2 and with a popular open-source OCR library -- Tesseract versions 3 and 4. The latest Tesseract engine uses deep learning models similar to ours.
Recognition is performed at the text line level. The ground truth layout structure is used to crop samples from document images.

Since it was shown that LSTMs learn an implicit language model\cite{Sabir2017}, we evaluate our system without external language models or lexicons, although their use can likely further increase accuracy.
By contrast, both examined commercial engines use language models and lexicons for English and German, and their settings have been chosen for best recognition accuracy.
We use the fast integer Tesseract 4 models\footnote{\url{https://github.com/tesseract-ocr/tessdata_fast}} because they demonstrate a comparable running time to the other examined methods.


Our data sources are summarized in~\tablename~{\ref{tab:datasources}}. We conduct experiments on the test documents with (\textit{Type-2}, \textit{Type-3}) and without (\textit{Type-1}) additional distortions (\ref{sec:data:augment}) applied prior to decoding. We explore two different scenarios for the degradations. In the first scenario, only geometrical transformations, morphological operations, blur, noise addition and downscaling are considered (\textit{Type-2}). This scenario corresponds to the typical case of printed document scans of varying quality. In the second scenario, all extracted text line images are additionally alpha-composited with a random background texture (\textit{Type-3}). Different texture sets are used for training and testing. Additionally, we randomly invert the image gray values. This scenario best corresponds to scene text recognition. Note that since the distortions are applied randomly, some images obtained by this procedure may end up nearly illegible, even for human readers.

We aggregate results from multiple experiments (every text line image is randomly distorted $\numprint{30}$ times) and report the average error values. \tablename~{\ref{tab:testdata}} summarizes our test datasets. We evaluate all methods on original and distorted text lines, containing $\numprint{222981}$ and $\numprint{10927830}$ characters, respectively.

{\setlength{\tabcolsep}{3pt}\renewcommand{\arraystretch}{1.0}
\newcommand\Tstrut{\rule{0pt}{2.2ex}}         
\begin{table}[!htbp]
\caption{Ground truth lengths of datasets used in our experiments.}
\label{tab:testdata}
\begin{center}
\begin{tabular}{ccccccc}
\toprule
\multicolumn{2}{c}{\textbf{Newspapers}} & \multicolumn{2}{c}{\textbf{Invoices}} & \multicolumn{3}{c}{\textbf{Synthetic documents}} \\
\cmidrule(lr{1.3em}){1-2}
\cmidrule(lr{0.9em}){3-4}
\cmidrule(l{0.9em}r{1.25em}){5-7}
Type-1 & Type-2 & Type-1 & Type-2 & Type-1 & Type-2 & Type-3\\
\midrule
$\numprint{66327}$ & $\numprint{1989810}$ & $\numprint{15374}$ & $\numprint{461220}$ & $\numprint{141280}$ & $\numprint{4238400}$ & $\numprint{4238400}$ \\
\bottomrule
\end{tabular}
\end{center}
\end{table}}

\tablename~\ref{tab:eval-acc} compares error rates of all examined OCR engines. We use the \textit{Levenshtein edit distance} metric\cite{levenshtein1966bcc} to measure the character error rate (CER). 
All of our models, unless otherwise stated, are fine-tuned with real data, use geometric text line normalization, and data augmentation methods (\ref{sec:data:augment}) except elastic distortions. 
The results show that our system outperforms all other methods in terms of recognition accuracy in all scenarios. A substantial difference can be primarily observed on distorted documents alpha-composited with background textures, where Tesseract and both commercial engines exhibit a very poor recognition performance. Noisy backgrounds hinder their ability to perform an adequate character segmentation. Although Tesseract 4 was trained on augmented synthetic data, we observe that it cannot properly deal with significantly distorted inputs. The established solutions have problems recognizing subscript and superscript symbols. Both commercial engines have great difficulties in handling fonts with different, alternating styles located on the same page.

\tablename~\ref{tab:top-errors} gives an insight into the most frequent errors (insertions, deletions, and substitutions) made by the best performing proposed and commercial methods on real versus synthetic data. All tested methods have the most difficulties in recognizing the exact number of whitespace characters due to non-uniform letter and word spacing (kerning, justified text) across documents. This problem is particularly visible on the manually corrected real documents, where a certain degree of ambiguity due to human judgment becomes apparent. The remaining errors for the hybrid models look reasonable and seem to be primarily focused on small or thin characters, which are indeed the ones most affected by distortions and background patterns. In contrast, ABBYY FineReader exhibits a clear tendency to insert spurious characters, especially for highly textured and distorted images.

{\setlength{\tabcolsep}{1.2pt}\renewcommand{\arraystretch}{0.92}
\newcommand\Tstrut{\rule{0pt}{3ex}}         
\newcommand\Bstrut{\rule[-0.9ex]{0pt}{0pt}}   
\begin{table}[!htbp]
\caption{Character error rates (\%) on the test datasets.}
\label{tab:eval-acc}
\begin{center}
\begin{tabular}{lrrrrrrr}
\toprule
& \multicolumn{2}{c}{\textbf{Newspapers}} & \multicolumn{2}{c}{\textbf{Invoices}} & \multicolumn{3}{c}{\textbf{Synthetic}}\\
\cmidrule(lr){2-3}
\cmidrule(lr){4-5}
\cmidrule(lr){6-8}
& Type-1 & Type-2 & Type-1 & Type-2 & Type-1 & Type-2 & Type-3 \\
\midrule
ABBYY FineReader                    & $0.63$ & $3.03$ & $2.65$ & $4.38$ & $6.64$ & $10.86$ & $19.27$ \\
OmniPage Capture                 & $0.31$ & $3.76$ & $2.61$ & $9.94$ & $7.62$ & $17.13$ & $58.43$ \\
Tesseract 3               & $1.15$ &$16.90$ & $6.11$ &$10.16$ & $11.79$ & $17.76$ & $37.00$ \\
Tesseract 4               & $1.14$ & $9.63$ & $4.53$ & $6.66$ & $8.70$ & $14.80$ & $35.91$ \\
\midrule
\OCRName{}$^{\mathrm{FCN}}$     & $0.16$ & $0.81$ & $1.66$ & $3.28$ & $1.03$ & $2.02$  & $3.25$ \\
\OCRName{}$^{\mathrm{Hybrid}}$     & $\bm{0.11}$ & $0.75$ & $\bm{1.63}$ & $\bm{2.33}$ & $0.62$ & $1.48$  & $2.84$ \\
\OCRName{}$^{\mathrm{Hybrid,nG}}$   & $0.13$ & $1.14$ & $2.86$ & $3.67$ & $\bm{0.46} $ & $\bm{1.31}$  & $4.50$ \\
\OCRName{}$^{\mathrm{Hybrid,Peep}}$     & $0.14$ & $\bm{0.69}$ & $1.85$ & $2.81$ & $0.53$ & $1.35$  & $\bm{2.43}$ \\
\OCRName{}$^{\mathrm{Hybrd,E}}$   & $\bm{0.11}$ & $0.73$ & $\bm{1.63}$ & $2.47$ & $0.65$ & $1.46$  & $2.70$ \\
\OCRName{}$^{\mathrm{Hybrid,nA}}$   & $0.14$ & $1.24$ & $2.49$ & $3.58$ & $0.48$ & $1.94$  & $7.54$ \\
\OCRName{}$^{\mathrm{Hybrid,S}}$   & $0.20$ & $0.96$ & $1.85$ & $3.05$ & $0.67$ & $1.58$  & $2.84$ \\
\OCRName{}$^{\mathrm{Hybrid,S,E}}$ & $0.20$ & $0.95$ & $1.76$ & $2.89$ & $0.69$ & $1.51$  & $2.69$ \\
\bottomrule
\multicolumn{8}{l}{$^{\mathrm{FCN}}$ denotes the fully convolutional model.}\Tstrut\\
\multicolumn{8}{l}{$^{\mathrm{Hybrid}}$ denotes the hybrid model.}\\
\multicolumn{8}{l}{$^{\mathrm{nG}}$ denotes that no geometric normalization was used.}\\
\multicolumn{8}{l}{$^{\mathrm{Peep}}$ denotes the use of peephole LSTM units.}\\
\multicolumn{8}{l}{$^{\mathrm{E}}$ denotes training with elastic distortions.}\\
\multicolumn{8}{l}{$^{\mathrm{nA}}$ denotes training without alpha compositing with background textures.}\\
\multicolumn{8}{l}{$^{\mathrm{S}}$ denotes training exclusively with synthetic data.}\\
\end{tabular}
\end{center}
\end{table}}

\begin{figure*}[!htbp]
\centering
\begin{tabular}{c}
\includegraphics[height=5.5cm]{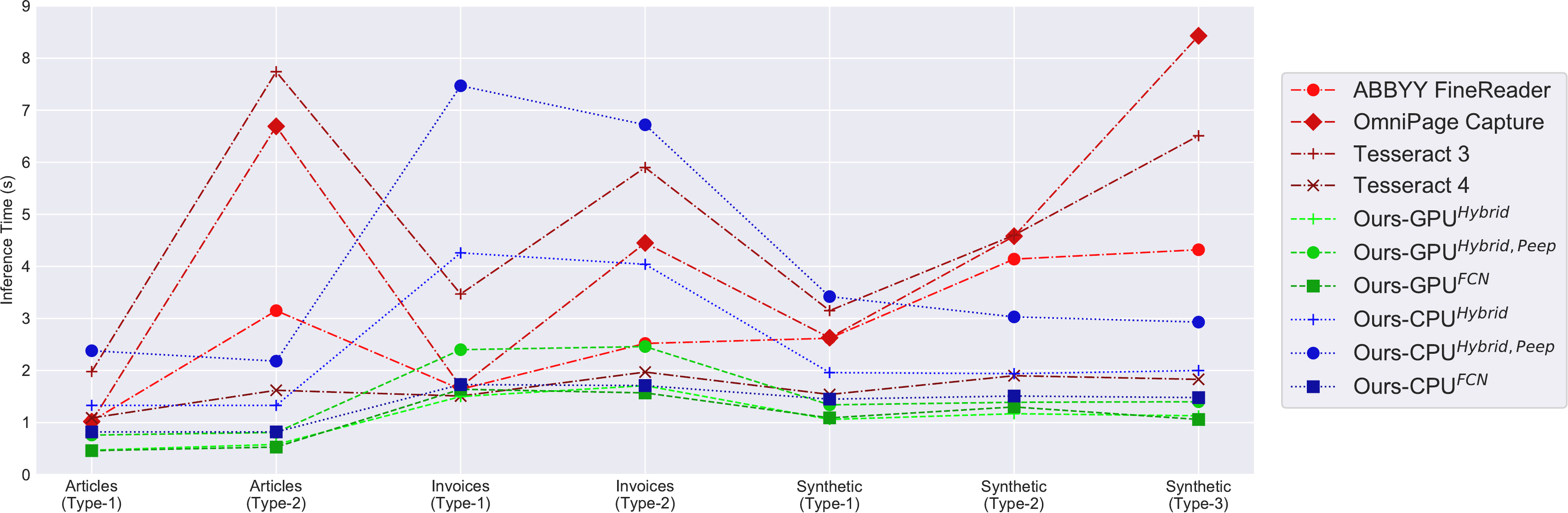}\\
\end{tabular}
\caption{Runtime comparison (in seconds) on the test datasets (per standard page with \numprint{1500} symbols). Values for the original documents (\textit{Type-1}) and CPU experiments are averaged over \numprint{10} and for all other experiments over \numprint{30} trials. All CPU experiments use a batch size of $4$ images, whereas the GPU runs use a batch of $48$ images. Note that data points from different datasets are connected solely to allow easier traceability of each engine.}
\label{fig:eval-speed}
\end{figure*}

{\setlength{\tabcolsep}{1.5pt}\renewcommand{\arraystretch}{1.0}
\newcommand\Tstrut{\rule{0pt}{2.0ex}}         
\newcommand\Bstrut{\rule[-0.9ex]{0pt}{0pt}}   
\begin{table}[!htbp]
\begin{center}
\caption{Top 10 most frequent errors for the best proposed- and commercial OCR engines. Note that the real data (left side) comprises both newspapers and invoices.}
\label{tab:top-errors}
\begin{subtable}[t]{0.48\columnwidth}
\centering
\begin{tabularx}{0.99\columnwidth}{Xr}
\toprule
Error type & \%\\
\midrule
Insertion of ' ' & $\numprint{25.77}\%$\\
Substitution 'l'$\rightarrow$'i' & $\numprint{4.43}\%$\\
Substitution '.'$\rightarrow$',' & $\numprint{3.56}\%$\\
Insertion of '.' & $\numprint{2.77}\%$ \\
Substitution 'i'$\rightarrow$'l'& $\numprint{2.67}\%$ \\
Insertion of '\_' & $\numprint{2.52}\%$\\
Substitution 'I'$\rightarrow$'l' & $\numprint{2.15}\%$ \\
Insertion of 't' & $\numprint{1.27}\%$ \\
Substitution 'o'$\rightarrow$'a' & $\numprint{1.21}\%$ \\
Substitution 'f'$\rightarrow$'t' & $\numprint{1.19}\%$ \\
\bottomrule
\end{tabularx}
\caption{\Tstrut\footnotesize \OCRName{}$^{\mathrm{Hybrid}}$ (real data)}
\end{subtable}
\hspace{\fill}
\begin{subtable}[t]{0.48\columnwidth}
\centering
\begin{tabularx}{0.99\columnwidth}{Xr}
\toprule
Error type & \%\\
\midrule
Insertion of ' ' & $\numprint{8.53}\%$ \\
Substitution '0'$\rightarrow$'O' & $\numprint{4.09}\%$ \\
Deletion of '.' & $\numprint{1.77}\%$ \\
Substitution 'O'$\rightarrow$'Ö'& $\numprint{1.62}\%$ \\
Deletion of '\_' & $\numprint{1.58}\%$ \\
Deletion of '-' & $\numprint{1.45}\%$ \\
Substitution 'I'$\rightarrow$'l' & $\numprint{1.44}\%$ \\
Substitution '.'$\rightarrow$',' & $\numprint{1.33}\%$ \\
Insertion of 'r' & $\numprint{1.00}\%$ \\
Substitution '©'$\rightarrow$'O' & $\numprint{0.97}\%$ \\
\bottomrule
\end{tabularx}
\caption{\Tstrut\footnotesize \OCRName{}$^{\mathrm{Hybrid,Peep}}$ (synthetic data)}
\end{subtable}
\begin{subtable}[t]{0.48\columnwidth}
\centering
\begin{tabularx}{0.99\columnwidth}{Xr}
\toprule
Error type & \%\\
\midrule
Insertion of ' ' & $\numprint{12.35}\%$\\
Insertion of '.' & $\numprint{5.70}\%$\\
Substitution ','$\rightarrow$'.' & $\numprint{2.95}\%$\\
Insertion of 'i' & $\numprint{2.70}\%$\\
Insertion of 'r' & $\numprint{2.09}\%$\\
Deletion of 'e' & $\numprint{1.83}\%$\\
Substitution 'c'$\rightarrow$'e' & $\numprint{1.77}\%$\\
Insertion of 'l' & $\numprint{1.63}\%$\\
Insertion of 'n  & $\numprint{1.41}\%$\\
Insertion of 't' & $\numprint{1.35}\%$\\
\bottomrule
\end{tabularx}
\caption{\footnotesize{ABBYY FineReader (real data)}}
\end{subtable}
\hspace{\fill}
\begin{subtable}[t]{0.48\columnwidth}
\centering
\begin{tabularx}{0.99\columnwidth}{Xr}
\toprule
Error type & \%\\
\midrule
Insertion of ' ' & $\numprint{9.27}\%$ \\
Insertion of 'i' & $\numprint{1.72}\%$ \\
Insertion of 'e' & $\numprint{1.55}\%$ \\
Insertion of 't' & $\numprint{1.33}\%$ \\
Insertion of 'r' & $\numprint{1.21}\%$ \\
Insertion of '.' & $\numprint{1.19}\%$ \\
Insertion of 'l' & $\numprint{1.18}\%$ \\
Insertion of '-' & $\numprint{1.07}\%$ \\
Insertion of 'n' & $\numprint{1.06}\%$ \\
Insertion of 'a' & $\numprint{0.96}\%$ \\
\bottomrule
\end{tabularx}
\caption{\footnotesize{ABBYY FineReader (synthetic data)}}
\end{subtable}
\end{center}
\end{table}
}

\figurename~\ref{fig:eval-speed} presents the runtime comparison.
Both commercial engines and Tesseract 3 work slowly for significantly distorted images. 
Apparently, they make use of certain computationally expensive image restoration techniques in order to be able to handle low-quality inputs.
Unsurprisingly, the GPU-accelerated models are fastest across the board. We discuss the runtime on CPU in~\ref{sec:ablation}.
All experiments were conducted on a workstation equipped with an Nvidia GeForce~GTX~745 graphics card and an Intel~Core~i7-6700 CPU.


\subsection{Ablation study}
\label{sec:ablation}
In this section, we analyze the impact of different model components on the recognition performance of our system.

\subsubsection{Fully convolutional vs. recurrent model}
The fully convolutional model achieves a slightly lower accuracy than the best recurrent variant. However, its inference time is significantly lower on the CPU. This clearly shows that convolutional layers are much more amenable to parallelization than recurrent units.


\subsubsection{Peephole connections}
The model that uses peephole LSTM cells and pools feature maps along the width dimension only once, exhibits a better recognition accuracy in the scene text scenario. This is not the case for typical document scans, where the peepholes do not seem to bring any additional accuracy gains compared to the vanilla LSTM model. The use of peephole connections does, however, add a significant runtime overhead in all cases.

\subsubsection{Alpha compositing with background textures~(\ref{sec:data:augment})}
We train one model without alpha compositing with background textures. The model exhibits significantly higher error rates, not only on samples with complicated backgrounds but also on those with significant distortions. This confirms our assumption that this augmentation technique has generally a positive effect on the robustness of neural OCR models.


\subsubsection{Geometric normalization~(\ref{sec:data:geom_norm})}
The model using no geometric normalization exhibits a drop in accuracy especially for images showing stronger distortions. This indicates that geometric normalization is indeed beneficial, but not indispensable. Apparently, max pooling and strided convolution operations provide enough translational invariance.

\subsubsection{Training only on synthetic data}
We train two models exclusively on synthetic training data. It obtains very competitive results, which indicates that using such a model together with proper data augmentation is sufficient for achieving a satisfactory recognition accuracy.

\subsubsection{Elastic distortions}
We found that non-linear distortions can further reduce the error rate of models, particularly those trained exclusively on synthetic data. Hence, this augmentation method is beneficial, especially in cases where annotated real data is not available or simply too difficult to produce. We also observe that although most of our models were trained without elastic distortions applied to training data, they can nonetheless deal with test data augmented with non-linear distortions. We attribute this to the fact that we used a substantial amount of fonts to generate our synthetic training data, achieving an adequate variation of text styles.

\section{Conclusions and future work}
\label{sec:conclusions}
In this paper, we described our general and efficient OCR solution. Experiments under different scenarios, on both real and synthetically-generated data, showed that both proposed architectures outperform leading commercial and open-source engines. In particular, we demonstrated an outstanding recognition accuracy on severely degraded inputs. 

The architecture of our system is universal, and can be used to recognize printed, handwritten or scene text. The training of models for other languages is straightforward. 
Via the proposed pipeline, deep neural network models can be trained using only text line-level annotations. This saves a considerable manual annotation effort, previously required for producing the character- or word level ground truth segmentations and the corresponding transcriptions.

A novel data augmentation technique, alpha compositing with background textures, is introduced and evaluated with respect to its effects on the overall recognition robustness. Our experiments showed that synthetic data is indeed a viable and scalable alternative to real data, provided that sufficiently diverse samples are generated by the data augmentation modules. 
The effect of different structural choices and data augmentation on recognition accuracy and inference time is experimentally investigated. Hybrid recognition architectures proved to be more accurate, but also considerably more computationally costly than purely convolutional approaches.

The importance of a solid data generation pipeline cannot be overstated. As such, future work will involve its continuous improvement and comparison with other notable efforts from the research community, e.g.,~\cite{Jaderberg2014}. 
We also plan to make the synthetic data used in our experiments publicly available.
We feel that fully-convolutional approaches, in particular, offer great potential for future improvement. The incorporation of recent advances, such as residual connections~\cite{He2016} and squeeze-and-excitation blocks~\cite{hu2018senet} into our general OCR architecture seems to be a promising direction. 

\section*{Acknowledgment}
This work was supported by the German Federal Ministry of Education and Research (BMBF) funded program KMU-innovativ in the project DeepER.



\bibliographystyle{IEEEtran}
\bibliography{IEEEabrv,ocr_icdar2019}

\end{document}